\documentclass[11pt]{article}
\usepackage{amsmath}
\usepackage{graphicx,psfrag,epsf}
\usepackage{enumerate}
\usepackage[super,sort&compress]{natbib}
\usepackage{xcolor}
\usepackage{tabularx}
\usepackage{adjustbox}
\usepackage{multirow}
\usepackage{subcaption}
\usepackage{comment}
\usepackage{colortbl}
\usepackage{booktabs}
\usepackage{array}
\usepackage{enumitem}
\usepackage{hyperref}
\usepackage{tikz}
\usetikzlibrary{positioning,fit,arrows.meta}
\usepackage{xcolor}
\hypersetup{
    colorlinks,
    linkcolor={red!50!black},
    citecolor={blue!50!black},
    urlcolor={blue!80!black}
}

\newcommand{\blind}{0}

\begin{document}

%\def\spacingset#1{\renewcommand{\baselinestretch}%
%{#1}\small\normalsize} \spacingset{1}

%%%%%%%%%%%%%%%%%%%%%%%%%%%%%%%%%%%%%%%%%%%%%%%%%%%%%%%%%%%%%%%%%%%%%%%%%%%%%%

%\author{Author 1\thanks{}\hspace{.2cm}
\if0\blind
{
  \title{\textbf{Keyphrase Generative Representation of Youth Crisis Conversations Beyond Static Taxonomies}}

  \author{
    Abeer Badawi\textsuperscript{1,2},
    Will Aitken\textsuperscript{3},
    Lydia Sequeira\textsuperscript{4},
    Jocelyn Rankin\textsuperscript{4}, \\
    Maia Norman\textsuperscript{2},
    Elham Dolatabadi\textsuperscript{1,2}
  }
  \date{
    \textsuperscript{1}York University, Canada
    \textsuperscript{2}Vector Institute, Canada 
    \textsuperscript{3}Electrical and Computer Engineering, Queen's University, Canada 
    \textsuperscript{4}Kids Help Phone, Canada \\
  }
  \begingroup
  \hypersetup{hidelinks}
  \maketitle
  \endgroup
} \fi

% OTHER TITLES: 
%From Static Taxonomies to Generative Keyphrases in Youth Crisis Text Services
% Keyphrase Generative Representation for Youth Crisis Conversations Beyond Static Labels
% Taxonomy Expansion and Keyphrase Generative Representation in Youth Crisis Text Services

\begin{abstract}

Crisis Responders (CRs) rapidly assess thousands of youth SMS conversations each year to identify mental health concerns and guide support. Yet youth distress is increasingly expressed through evolving and context-specific language that often does not fit fixed-label taxonomies. This work analyzed 703,975 de-identified Kids Help Phone conversations (2018–2023) and expanded KHP's 19-label issue taxonomy into a 39-label hierarchical schema. We then introduce Keyphrase Generative Representation (KGR), a constrained LLM generating concise, conversation-specific keyphrases, evaluated across 129 conversations and 387 expert annotations. The expanded taxonomy achieved expert-consensus reliability, with an accuracy of 0.96, and expert review found that 81\% of keyphrases accurately reflected content and 74\% improved clarity. KGR surfaced identity-linked themes absent from the fixed taxonomy, including immigration problems and caregiver burden, and supported a topic-retrieval workflow that outperformed the manual analyst process by 0.45 in accuracy. KGR marks a shift toward hybrid, interpretable generative representations that extend crisis response beyond static taxonomies to surface emerging and culturally grounded patterns of youth distress.
\end{abstract}

\section{Introduction}

\label{sec:intro}
%Globally, 1 in 7 adolescents experiences a mental health disorder, accounting for 15\% of the disease burden in this age group. However, many remain unrecognized and untreated, and among people aged 15 to 29 years, suicide ranks third as the leading cause of death, underscoring persistent gaps in prevention and access to effective care 

Adolescent mental health challenges represent a largely hidden epidemic worldwide, with many conditions remaining unrecognized and untreated. Estimates suggest that 1 in 7 adolescents experience a mental health disorder, and among people aged 15 to 29 years, suicide ranks as the third leading cause of death, underscoring persistent gaps in prevention and access to effective care \cite{WHO2021}. Despite increasing demand for care, access to timely mental health support remains limited, particularly for youth experiencing acute distress. As a result, text-based crisis support services have become an increasingly important point of contact for young people seeking immediate and accessible support \cite{Patel2021}. In these settings, frontline Crisis Responders (CRs) must interpret unstructured, emotionally dense conversations to determine whether escalation, referral, or follow-up is warranted. Beyond individual interactions, crisis services also depend on post-conversation representations to support supervision, quality assurance, triage auditing, longitudinal reporting, and resource allocation. Consequently, how conversations are summarized and represented after they occur directly shapes how organizations monitor emerging needs, evaluate service delivery, and identify patterns of distress across populations \cite{Wiens2020, khp,turkington2020why, Ali2025MentalHealthAI,rose2007labels, owen2024ai}.

%However, when post-conversation representations fail to surface contextually and culturally specific expressions of distress, emerging patterns of need may remain undetected at scale. This limitation is particularly consequential for equity monitoring and service oversight, because processes such as triage audit, escalation review, and longitudinal reporting rely on how conversations are summarized after they occur \cite{turkington2020why, ali2025artificial}. When complex interactions are compressed into a narrow set of predefined issue labels, decisions may be shaped by incomplete information, contributing to delayed escalation, reduced sensitivity to emerging risks, and limited visibility into changing patterns of youth distress. This concern is especially salient in youth mental health contexts, where language describing distress is highly variable and culturally grounded \cite{rose2007labels, owen2024ai}.

To support these workflows at scale, most crisis services rely on fixed issue taxonomies that map conversations to predefined categories such as anxiety, depression, self-harm, or family conflict. These taxonomies provide consistency, auditability, and a shared operational language, that enables automated issue identification through machine learning models\cite{faiirv1-2024, khp}. Prior work has shown that multi-label classification models trained on such taxonomies can achieve strong agreement with expert annotations for established concerns. However, performance within a fixed label space does not guarantee that the representation captures the contextual information required frontline, supervisory, and organizational decision-making \cite{Patel2021, WHO2021, CIHI2024}.

This limitation is particularly important in youth mental health contexts, where distress is often expressed through culturally grounded, rapidly evolving, and context-dependent language. When complex conversations are compressed into a limited set of predefined labels, important contextual signals may remain invisible within organizational systems. Such representational gaps may influence escalation review, equity monitoring, and institutional understanding of emerging youth mental health needs. Expanding taxonomies may improve coverage, but maintaining exhaustively updated label systems becomes difficult as language and sociocultural contexts evolve over time.

Recent advances in large language models (LLMs) have introduced generative approaches that can surface salient themes directly from text \citep{suicide-risk-model-2021, first-time-suicide-2021, reddit-suicide-2022, treatment-efficacy-2020, efficacy-survey-2021, alliance-use-case-2020, sentencebert-alliance-2023, mhi-2023}. More broadly, generative AI may enable systems to capture nuanced or emerging experiences that are difficult to encode within static taxonomies \cite{zhang2025generative, ji2023rethinking}. However, unconstrained generative outputs also introduce important concerns related to hallucination, variable specificity, fairness, and reduced transparency \cite{Patel2021, ihealth-ethics-2022, ai-gone-mental-2020}. As a result, it remains unclear how generative language models can be integrated into crisis service analytics while preserving reliability, auditability, and operational consistency required for real-world deployment. %These considerations underscore the need for generative methods that operate within controlled and auditable representational frameworks. These issues are particularly consequential in crisis services, where representations inform supervision, quality assurance, and organizational decision-making \cite{Patel2021, ihealth-ethics-2022, ai-gone-mental-2020}.

These limitations motivate a reframing of issue identification as a process that is anchored to, but not exhausted by, static labels. To address this gap, we investigate whether crisis conversations can be represented through a hybrid framework that preserves the reliability of structured taxonomies while allowing contextual signals of distress to surface through constrained generative methods. This approach preserves the operational value of taxonomies while allowing meaningful context to surface in a structured manner. Generative methods, when appropriately constrained, may therefore function as complementary layers that enrich post-conversation representations without displacing established classification systems \cite{ji2023rethinking, owen2024ai}.

In this study, we evaluate this hybrid approach using real-world data from Kids Help Phone, a national youth crisis service in Canada \cite{khp}. We first examine whether expanding a legacy 19-label taxonomy \cite{faiirv1-2024} to a 39-label set improves expert agreement and representational coverage. We then introduce a constrained generative representation approach, Keyphrase Generative Representation (KGR), that augments static issue labels with concise, conversation-specific keyphrases generated by an LLM. Using 703,975 real-world crisis conversations and expert evaluation of 387 annotations, we assess the alignment of KGR outputs with clinical judgment, characterize the types of information revealed beyond static labels, and examine the potential of this approach to support interpretability and sense-making in crisis service operations. By grounding generative representations within established taxonomic frameworks and evaluating them through expert review, this work explores a pragmatic pathway for integrating generative AI into mental health analytics while maintaining the reliability and transparency required for real-world deployment \cite{Torous2025DigitalMentalHealth, Patel2021}. This study contributes three advances to crisis analytics research:

\begin{itemize}
     \item We evaluate a national-scale dataset of 703{,}975 de-identified youth crisis conversations and show that expanding the existing 19-label issue taxonomy to a 39-label hierarchical schema improves topic coverage and structured issue identification, enabling broader representation of youth mental health, identity-related, and safety concerns.
 
     \item We introduce Keyphrase Generative Representation (KGR), a constrained LLM-based framework that augments standard issue labels with concise, non-diagnostic, conversation-specific keyphrases. Expert evaluation showed that 81\% of generated keyphrases accurately reflected conversation content, while also revealing new contextual and identity-linked patterns beyond fixed labels.

    \item We provide expert and operational evidence that KGR surfaces emerging, culturally grounded concerns beyond static taxonomies---including immigration-related challenges, caregiver burden, Indigenous community contexts, and microaggressions---and that KGR-based topic retrieval outperformed the existing manual analyst workflow at Kids Help Phone with a 45-percentage-point improvement in accuracy (70\% for KGR vs.\ 25\% for manual retrieval on bullying-related conversations
\end{itemize}

\section{Methods}
\label{sec:meth}

This section describes the dataset, representation framework, and evaluation methodology used in this study. We investigate two complementary approaches for representing youth crisis conversations at scale: (1) structured issue identification through taxonomy-based multi-label classification, and (2) KGR, a constrained generative framework designed to augment static labels with concise, context-sensitive representations of conversation content.

\subsection{Dataset Description}
We analyze real-world text-based crisis support conversations between service users and trained CRs at KHP \cite{khp}. KHP is a national, 24/7, free, confidential, and multilingual e-mental health service in Canada, providing support through calls, SMS, and browser-based text platforms \cite{Mughal2024_Part1, Mughal2024_Part2}. This study was approved by the Research Ethics Board at York University (certificate no. 2025-357) and conducted under a data use agreement with KHP and Vector Institute \cite{vector_institute_website}. All data were de-identified prior to analysis in accordance with KHP's privacy policy and consent procedures. 

As part of routine frontline workflows, volunteer CRs document one or more issue labels at the end of each conversation to support service monitoring, quality improvement, and research. Because conversations often involve multiple, overlapping issues, this documentation process is time-consuming and challenging \cite{Gould2022, faiirv1-2024}. In addition to message content, the dataset includes conversational metadata such as timestamps, sentiment labels, conversation identifiers, and coarse-grained location information. The final dataset comprises 703{,}975 unique conversations collected between 2018 and 2023, with individual conversations typically ranging from 500 to 3{,}000 tokens. For modeling, we concatenated all conversations into a single text sequence, prefixing each message with the corresponding speaker label: CR, service user, or, in rare cases, unknown speaker.

\begin{figure*}
    \centering
    \includegraphics[width=\textwidth, height = 9cm]{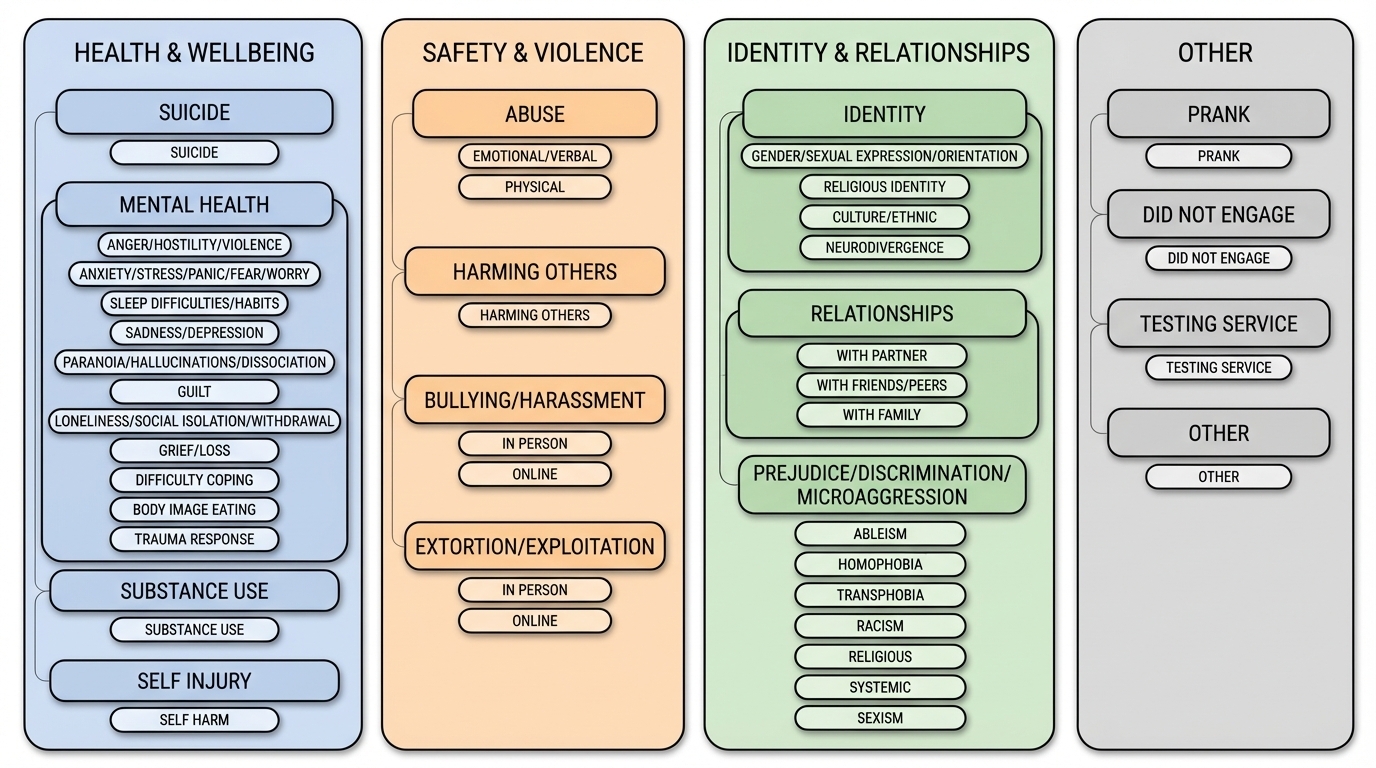}
    \caption{Hierarchical taxonomy of conversational tags used in our annotation framework. The schema contains four top-level categories (Health \& Wellbeing, Safety \& Violence, Identity \& Relationships, and Other) and 39 labels organized by subcategory.}
    \label{fig:tag_taxonomy}
\end{figure*}

As part of standard operational procedures, crisis responders (CRs) assign one or more issue labels following each interaction, using a taxonomy comprising 19 categories. This labeling process supports clinical supervision, quality assurance, and organizational reporting. In the present work, we expand this taxonomy from 19 to 39 labels to achieve more comprehensive and fine-grained coverage of mental health, safety, identity, and engagement-related concerns, as illustrated in Figure~\ref{fig:tag_taxonomy}. The expanded label set was developed through an iterative consultation process with clinical experts who also serve as frontline CRs and Supervisors at KHP, and who identified novel and emerging themes that have become increasingly salient in conversations over the past year.

\subsection{Representational Strategies}

\subsubsection{Multi-Label Classification}
\label{sec:mlc}

We expand the legacy 19-label issue taxonomy from previous work \cite{faiirv1-2024} to a 39-label hierarchical schema spanning four domains: mental health, safety, identity, and engagement, as shown in Figure~\ref{fig:tag_taxonomy}. Expansion was performed through iterative consultation with trained CRs to incorporate emerging and identity-linked themes observed in recent conversations.

Multi-label classification was implemented using the Llama~3~8B generative-AI model \cite{Meta2024Llama3} in a zero-shot configuration. The model, conditioned on the full conversation transcript and the expanded taxonomy, was prompted to assign all supported issue labels, with no fixed output cardinality. This taxonomy expansion served as a structured baseline for assessing the extent to which increased label granularity alone improves expert agreement and representational coverage.

The model produces a set of labels representing its assessment of the salient issues discussed (e.g., a bullying-related conversation may yield labels related to emotional abuse, relationships, anxiety or stress, guilt, and depression). Predicted labels were post-processed using semantic similarity–based scoring \cite{chandrasekaran2020evolution} to enable flexible alignment between model outputs and reference labels. While the expanded taxonomy improved coverage, its static structure motivated a generative approach to capture more details and emergent signals.

%\paragraph{Static Label Inference}
%We developed a large language model–based multi-label classification system to identify clinically and operationally relevant issues within youth crisis support conversations from KHP. The system maps free-text dialogue to a structured taxonomy of 39 fine-grained issue labels spanning mental health, safety, identity, and engagement domains as shown in Figure~\ref{fig:tag_taxonomy}. This design reflects the clinical reality that crisis conversations frequently involve multiple, overlapping concerns rather than a single dominant issue. 

%At inference time, the model conditions on the full conversation transcript, responder-identified topic cues, and the expanded label taxonomy. Classification is framed as a holistic semantic inference task and infer all supported labels from the conversational context without imposing a fixed output cardinality. We instantiate the classifier using Llama~3~8B \cite{Meta2024Llama3} in a zero-shot configuration, guided by a task-specific system prompt that instructs the model to assign all relevant labels. 

%\subsection{Multi-Label Classification Performance Evaluation}

\subsubsection{Keyphrase Generative Representation (KGR)}
\label{sec:key}

To overcome the limitations of static taxonomies, we introduce KGR, a novel generative approach that captures nuanced, context-rich signals that may not map clearly to the predefined issue labels. Rather than forcing conversations into fixed categories, KGR produces concise, conversation-specific descriptors that surface emerging themes, contextual factors, and identity-linked experiences present in youth crisis interactions.

\begin{figure*}[!tbp]
 \centering
  \subfloat{\includegraphics[width=16cm, height=7cm]{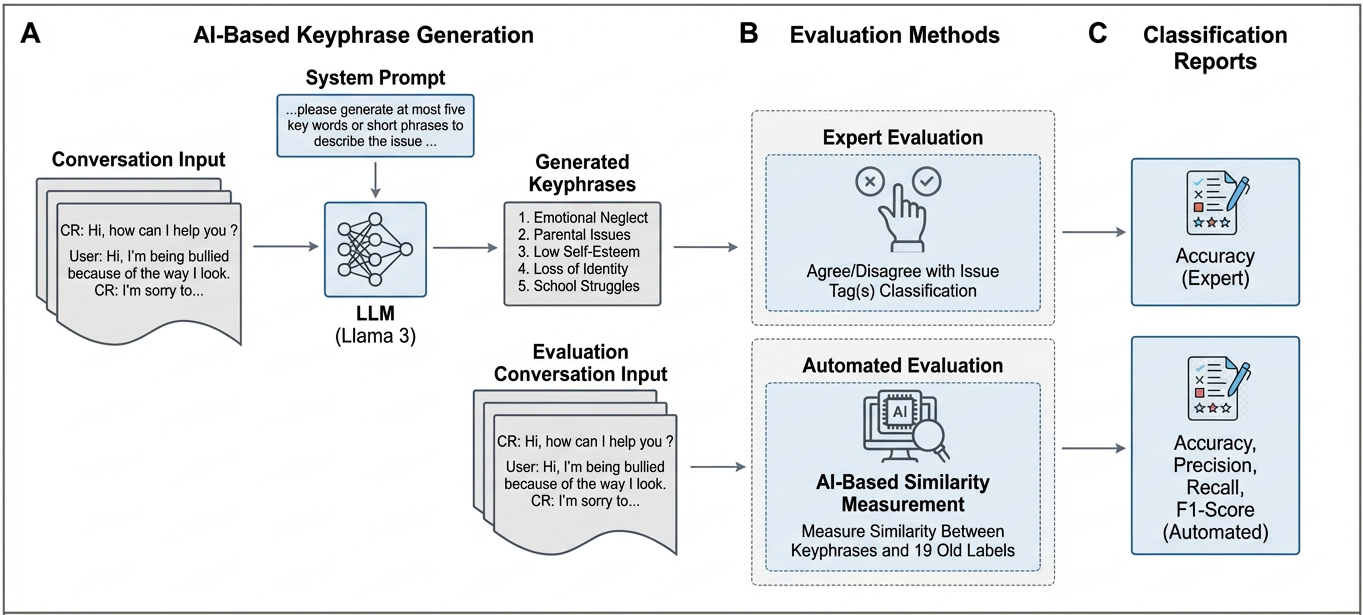}}
 \caption{The System architecture of the proposed Keyphrase Generative Representation (KGR) approach. The model takes a crisis conversation as input and produces a small set of concise keyphrases describing the main issues discussed. These generated representations are subsequently evaluated by experts and by automated similarity matching to legacy labels, supporting interpretable crisis analytics beyond static taxonomies.}
 \label{fig:systemkgr}
\end{figure*}

The diagram in Figure \ref{fig:systemkgr} illustrates the overall architecture of the proposed keyphrase generation and evaluation pipeline.

\textbf{(A) AI-Based Keyphrase Generation:} The process begins with conversation inputs between a user and a crisis responder (CR), which serve as the raw textual data for the system. These conversations are provided to the Llama~3 model through a system prompt instructing the model to generate up to five concise keyphrases representing the main issues discussed in the interaction. The output consists of 5 generated keyphrases summarizing the core themes of the conversation.

\textbf{(B) Evaluation Methods:} The generated keyphrases are then assessed using two complementary evaluation strategies. First, an expert evaluation is conducted in which domain experts manually determine whether the generated keyphrases appropriately match the issue classifications. Second, an automated evaluation computes AI-based similarity measurements between the generated keyphrases and the original 19 issue labels.

\textbf{(C) Classification Reports:} Finally, the evaluation results are summarized into classification reports. The expert evaluation produces accuracy-based assessments, while the automated evaluation reports quantitative metrics including accuracy, precision, recall, and F1-score. To ensure interpretability and operational safety, generation is restricted to brief, non-diagnostic phrases aligned with the language used by youth in crisis conversations. 

The resulting keyphrases serve as a structured generative layer that complements taxonomy-based classification by providing richer contextual signals while preserving concise representation. Lastly, the evaluation procedures for KGR followed two approaches: expert assessment by CRs, and automated alignment with existing issue labels, as explained in the next section.

%Discussion: This step ensures that the system is reliable and aligned with expert insights. 

%

%The expert human evaluation is conducted by reviewing the generated keyphrases and indicates agreement or disagreement with whether they accurately reflect the issues discussed in the conversation. In addition, an automated evaluation calculates a similarity score between the generated keyphrases and 19 pre-existing labels from our previous model \cite{faiirv1-2024} \cite{khp}, indicating how well the new keyphrases align with known classifications. The last component is a classification report that shows how often the model correctly identified keyphrase issues in the conversation, as evaluated by experts. This step ensures that the system is reliable and aligned with expert insights. For automated evaluation, the report also includes precision, recall, and F1-score, along with accuracy, to assess performance \cite{hicks2022evaluation}.

\subsection{Performance Evaluation}

\paragraph{Expert-based Evaluation}
All expert evaluations were conducted by CRs at KHP with advanced training in youth mental health crisis intervention and experience in supervision and quality assurance roles.

\paragraph{Decision-relevant evaluation framework}
The multi-label classification and KGR strategies were evaluated by trained experts using a structured rubric designed to reflect operational use in crisis services. For each conversation, evaluators indicated whether the representation covered all relevant topics (Yes/No), rated overall effectiveness across four predefined dimensions (reflection of lived experiences, identification of emerging topics, capture of subtle details, and usefulness for understanding the conversation) using a three-point scale (\textit{Agree, Neutral, Disagree}), and compared the representation with the prior 19-label \texttt{FAIIR V1} system \cite{faiirv1-2024,khp} using a five-point comparative scale. For KGR specifically, experts also evaluated each generated keyphrase for applicability to the conversation (Yes/No), and agreement ratios were computed from these evaluations.

\paragraph{Label-level Agreement for Static Labels}
For taxonomy-based classification, a stratified subset of conversations was independently annotated by three experts, who each selected and ranked up to five labels from the 39-label taxonomy. To account for inter-annotator variability, three aggregation schemes were examined: \textit{any} (selected by at least one annotator), \textit{Top-2 majority} (within the top two labels for at least two annotators), and \textit{Top-2 consensus} (within the top two labels for all three annotators). Model predictions were compared against these reference labels to compute accuracy, precision, recall, F1-score, and Area Under the Receiver Operating Characteristic curve (AUROC) \cite{hicks2022evaluation}.

\paragraph{Automated Evaluation of KGR}

For KGR, automated evaluation required mapping generated keyphrases to the legacy issue taxonomy to assess whether generative outputs captured information comparable to structured classification. Alignment strategies included \textit{exact matching}, \textit{exact matching with sublabels}, \textit{embedding-based similarity} using \texttt{all-MiniLM-L6-v2}, and variants incorporating sublabels and class-specific similarity thresholds. Exact matching methods checked whether generated keyphrases contained label names or related sublabels (e.g., \textit{Addiction} for \textit{Substance Abuse}). Semantic strategies embedded both keyphrases and labels using the \texttt{all-MiniLM-L6-v2}\footnote{\url{https://huggingface.co/sentence-transformers/all-MiniLM-L6-v2}} model and assigned labels when cosine similarity exceeded predefined thresholds. Additional variants incorporated sublabel expansion and per-class threshold tuning to balance precision and recall. Performance was measured using sample-averaged precision, recall, and F1-score consistent with \texttt{FAIIR V1} reporting.

\subsection{Rationale for Model Selection}
We required a model capable of both multi-label classification over an expanded 39-topic taxonomy and generating keyphrases aligned with youth language. Llama 3 was selected as the optimal self-hostable large language model due to three primary advantages \cite{grattafiori2024llama3}. Llama 3 offers significantly greater modeling capacity than Longformer model (as was used in our previous study) \cite{faiirv1-2024} \cite{khp} , enabling a more nuanced understanding of sentiment, subtext, and evolving youth vernacular \cite{singhal2023largelanguage}. Its generative abilities enable the system to produce concise, context-rich keyphrases, differentiate between major and minor issues, propose emergent subtopics, and generate interpretable rationales, aligning with prior work showing that adapted large language models can outperform medical experts in clinical text summarization and support clinically useful, task-specific summaries \cite{vanveen2024clinical,bednarczyk2025scientific}. Finally, Llama 3 satisfies KHP’s privacy and infrastructure requirements, as it can be fully deployed on-premises. Quantized variants allow for real-time inference while maintaining architectural transparency, model controllability, and alignment with privacy-preserving deployment practices reported for locally hosted clinical LLM pipelines \cite{wiest2024llmaix}.

\subsection{Chain-of-Thought Prompting and Insight Generation}
\label{sec:methods_topic_retrieval}

Chain-of-Thought (CoT) zero-shot prompting~\citep{zero-shot-cot-2022} improves LLM reasoning by eliciting intermediate steps, mimics human problem-solving, and enables models to break complex problems into smaller parts without prior training. In this work, CoT is used to surface the in-conversation evidence supporting a generated keyphrase or topic match, providing interpretable justifications that frontline staff and supervisors can audit.

\paragraph{Insight Generation.} A primary motivator behind open-ended keyphrase generation was to enable the identification of new issues and topics not covered by a predefined taxonomy. Kids Help Phone (KHP) regularly searches its conversation database for instances of report- or funding-specific topics (e.g.\ Climate Anxiety) that do not map cleanly to existing labels. Our system measures cosine similarity between a proposed topic and the keyphrases of each conversation in a selected time period; above-threshold matches can then be passed to Llama~3 with a CoT prompt to return verbatim supporting excerpts. Automating this procedure substantially reduces the burden on report authors, allowing them to focus on writing rather than data curation.

\paragraph{Evaluation.} For evaluation, a KHP data analyst manually matched conversations to two representative topics (\textit{bullying} and \textit{body image}) using keyword search over the same conversation pool, blinded to system outputs. Topic-relevance ground truth was derived from the expert evaluations, with trained Crisis Responders judging whether each conversation reflected the target topic. We compared both methods using accuracy, precision, recall, and F1-score.

\section{Results}
%Expansion from 19 to 39 labels served as a structured baseline to quantify the extent to which increased label granularity improves expert agreement and representational coverage, providing a reference for evaluating whether generative augmentation adds information beyond static classification.

%We report results across three stages: (i) characteristics of the expert-annotated evaluation sample, (ii) performance of taxonomy-based multi-label classification under alternative expert aggregation schemes, and (iii) evaluation of the proposed KGR framework using automated and expert assessments. Results are presented with emphasis on alignment with expert judgment, robustness across issue categories, and qualitative evidence of contextual and nuanced understanding. %Specifically, Phase~1 reports results for multi-label classification using the expanded 39-label taxonomy, and Phase~2 reports results for KGR using automated semantic benchmarks and expert human assessment.

We evaluated whether expanding structured taxonomies and augmenting them with constrained generative representations, KGR, improves the representation of youth crisis conversations at scale. Results are presented across three stages: (1) characterization of the expert-annotated evaluation subset, (2) evaluation of taxonomy-based multi-label classification under alternative expert aggregation schemes, and (3) assessment of KGR using automated and expert evaluation procedures. Across analyses, we focus on alignment with expert judgment, representational coverage, and the ability to surface contextual signals beyond fixed categorical labels.

\subsection{Evaluation Sample Characteristics}

\begin{figure*}
 \centering
 \resizebox{1\linewidth}{!}{\includegraphics{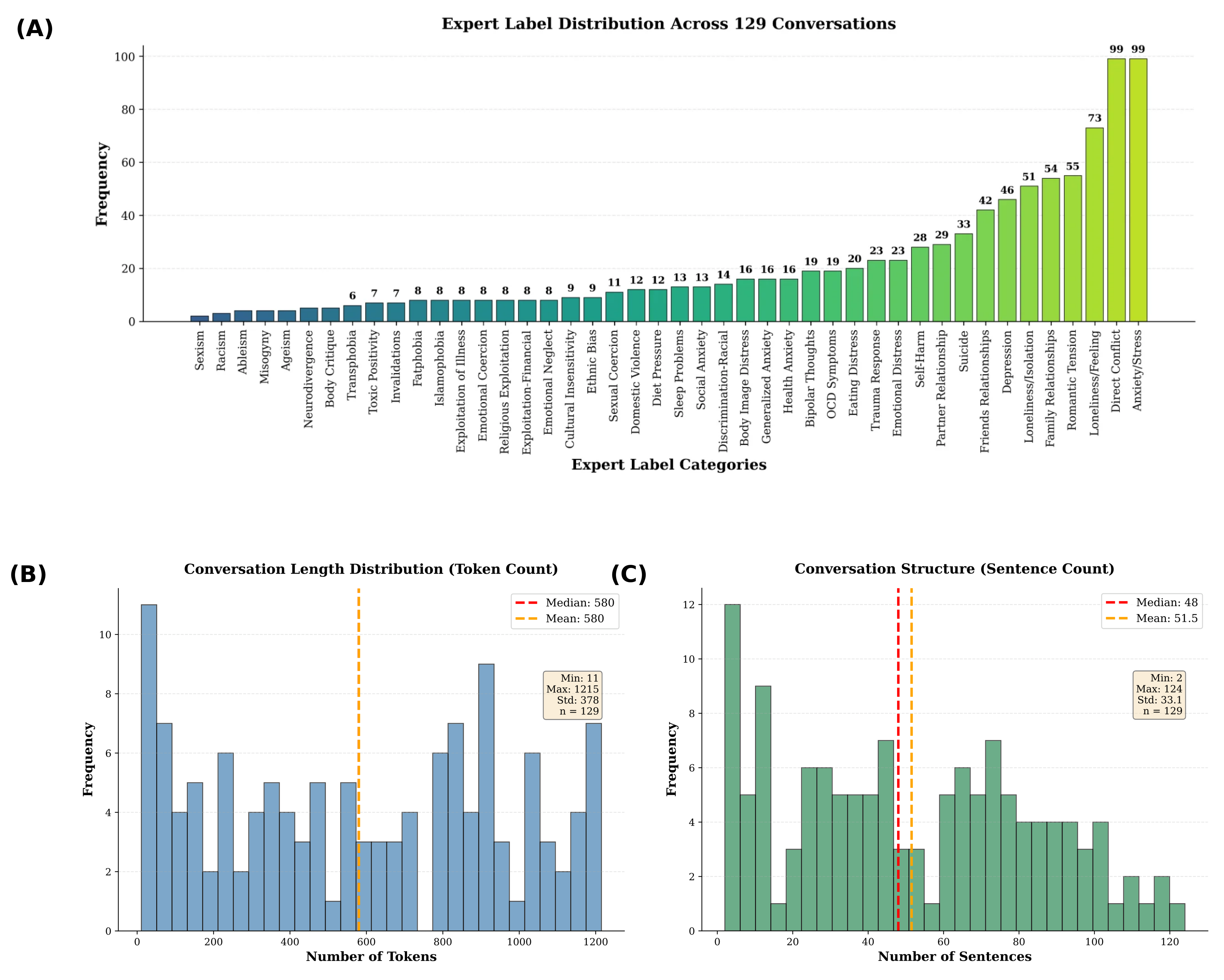}}
 \caption{Label distribution of 129 conversations annotated for the survey. (A) Expert-label distribution shows the frequency of mental health issue categories. (B) Conversation length distribution measured by the token count. (C) Conversation structure is measured by the sentence count.}
 \label{fig:label_dist}
\end{figure*}

The expert evaluation subset comprised 129 crisis support conversations sampled from the full dataset of 703{,}975 interactions. Each conversation was independently annotated by three trained CRs, resulting in 387 expert annotations. Figure~\ref{fig:label_dist}(A) illustrates the distributions of expert-assigned issue labels across the 129 conversations reflect the multi-label and heterogeneous nature of youth crisis conversations. Frequently-assigned categories included anxiety/stress, loneliness, relationship concerns, and depression. Less frequent but clinically salient labels captured discrimination-related experiences, neurodivergence, identity-related stressors, and safety concerns. This distribution underscores the operational challenge of representing both common and low-prevalence but high-importance issues within a fixed taxonomy.

Panels (B) and (C) characterize the structural properties of the conversations themselves. As shown in Figure~\ref{fig:label_dist}(B), conversation length varies substantially, ranging from 11 to 1{,}215 tokens with a median of 580 tokens. The broad range indicates that crisis interactions may range from brief exchanges to extended dialogues, depending on the situation's complexity. Figure~\ref{fig:label_dist}(C) shows a similar pattern for sentence counts, with conversations spanning from 2 to 124 sentences and a median of 48 sentences.

%Conversations ranged from 11 to 1{,}215 tokens, with a median length of 580 tokens, indicating substantial variation in conversational depth within the evaluation set. Panel~C shows the distribution of conversation length, measured in sentences, with counts ranging from 2 to 124 and a median of 48 per conversation. These distributions highlight the diversity in both topical content and structural complexity represented in the expert-annotated sample.

\subsection{Phase 1: Taxonomy-Based Multi-Label Classification}
\begin{figure*}[!tbp]
\centering

\subfloat[Multi-label classification results comparing three expert-label assignment schemes: Top-2 Majority (label in the top two for at least 2 annotators), Top-2 Consensus (label in the top two for all 3 annotators), and Any (label selected by any annotator).]{%
\includegraphics[width=0.85\textwidth]{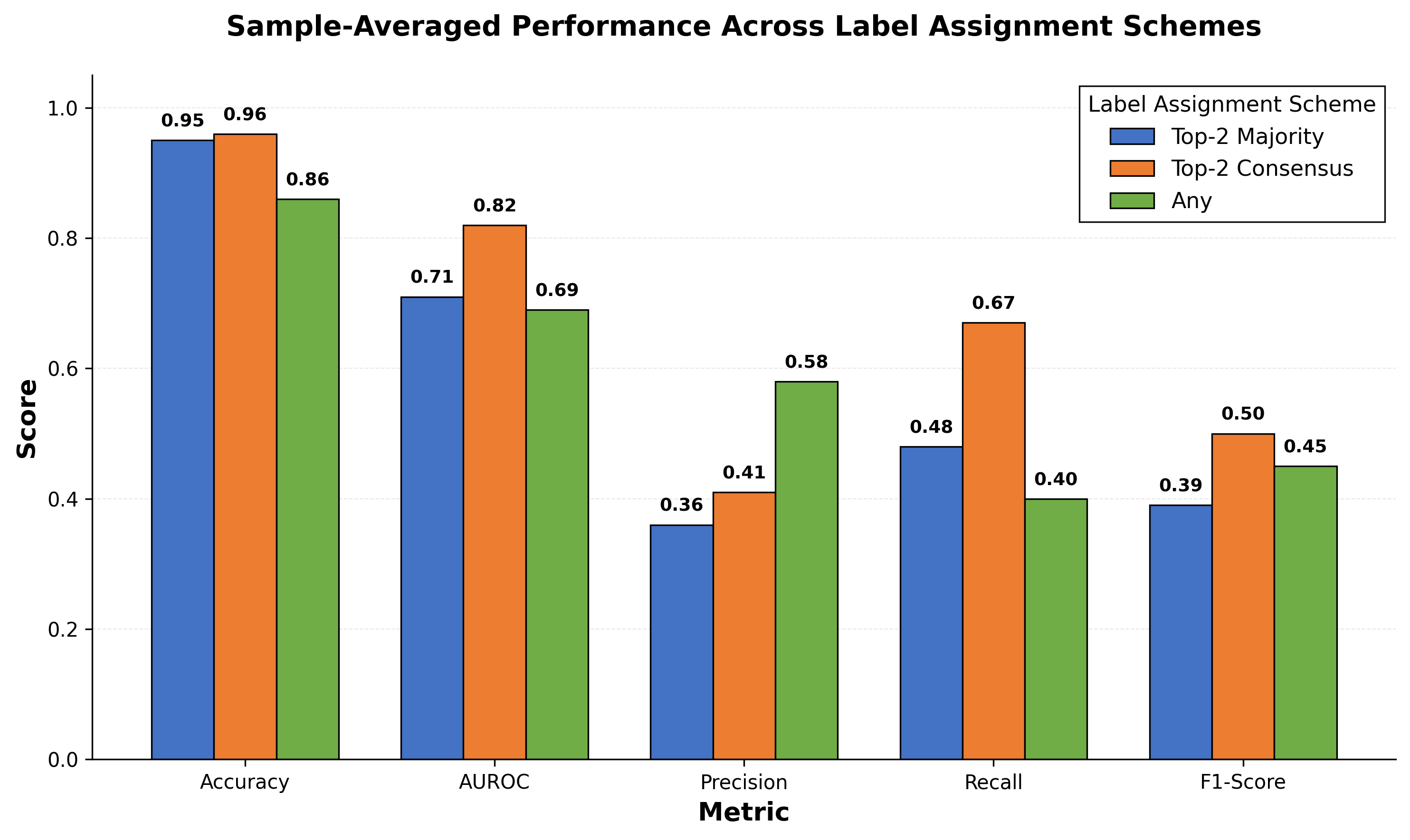}%
\label{fig:results_class}}

\vspace{0.5cm}

\subfloat[F1-scores across the 39 issue tags for multi-label classification, evaluated under the Top-2 Consensus expert label aggregation scheme.]{%
\includegraphics[width=0.85\textwidth]{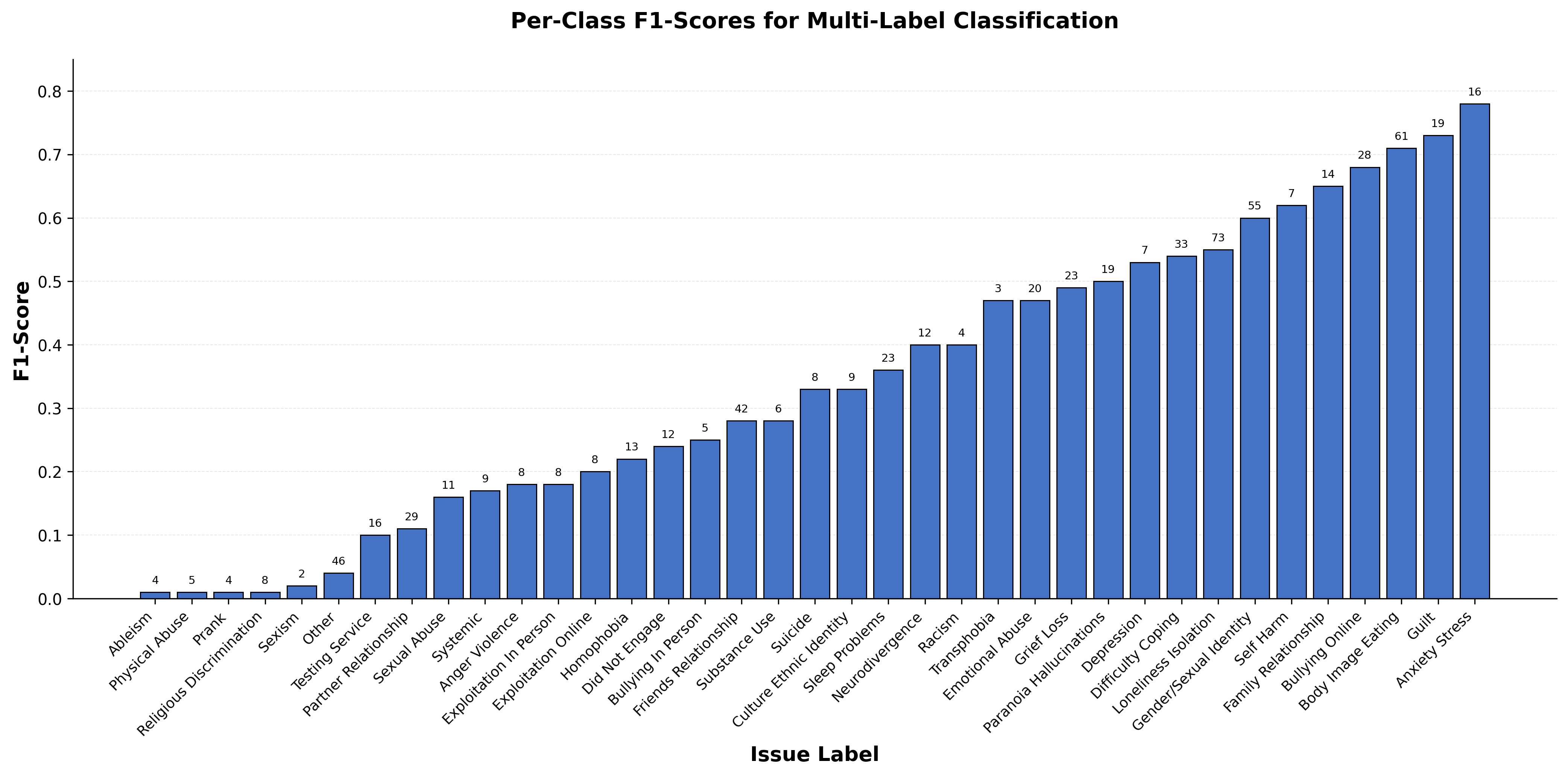}%
\label{fig:flscore}}

\caption{Evaluation of multi-label classification performance. (a) Comparison of expert label aggregation schemes used for model evaluation. (b) Per-class F1-scores across the 39 issue tags under the Top-2 Consensus scheme.}
\label{fig:classification_results}
\end{figure*}

We first evaluated whether expanding the issue taxonomy from 19 to 39 labels improves the reliability and representational coverage of structured issue identification. Conversations were sampled from the full dataset using a stratified strategy that prioritized challenging, ambiguous, and multi-topic cases. Figure~\ref{fig:results_class} compares model performance across the three predefined expert label aggregation schemes. The "Top-2 consensus" condition achieved the strongest overall performance, with an accuracy of 0.96 and an AUROC of 0.82, exceeding both "Top-2 majority" (accuracy 0.95, AUROC 0.71) and Any (accuracy 0.86, AUROC 0.69). 

%—Top-2 Majority, Top-2 Consensus, and Any—using accuracy, AUROC, precision, recall, and F1-score as evaluation metrics. Across schemes, the Top-2 Consensus condition achieved the strongest overall performance, with an accuracy of 0.96 and an AUROC of 0.82, exceeding both Top-2 Majority (accuracy 0.95, AUROC 0.71) and Any (accuracy 0.86, AUROC 0.69). 

Precision was highest under the "any" scheme (0.58), reflecting broader label inclusion, whereas recall was highest under "Top-2 consensus" (0.67), followed by "Top-2 majority" (0.48) and Any (0.40). Per-class F1-scores (Figure~\ref{fig:flscore}) demonstrate consistent improvements under the consensus condition across the 39 issue tags. The consensus-based evaluation provided the best balance between sensitivity and overall discrimination for static issue identification.

\subsection{Phase 2: Generative Augmentation with KGR}
\label{KPG}

As an initial step, we conducted an automated evaluation to assess whether KGR aligned with known issue categories prior to human review. Using the 19 \texttt{FAIIR V1} survey classes as proxy references \cite{faiirv1-2024}, we compared alignment against a random baseline (Precision = 0.16, Recall = 0.55, F1-score = 0.23) as shown in Table \ref{tab:kp_auto_results}. "Exact matching" achieved moderate alignment (Precision = 0.45, Recall = 0.24, F1-score = 0.30), while incorporating "sublabels" significantly improved performance (Precision = 0.64, Recall = 0.42, F1-score = 0.47). Embedding-based "similarity" methods showed comparable performance, with threshold-based methods achieving F1-score = 0.40 to 0.49. Methods combining sublabels with tuned thresholds produced results similar to those with sublabels alone (Precision = 0.57, Recall = 0.51, F1-score = 0.49). These results demonstrate that generated keyphrases reliably align with expert-defined categories, supporting progression to expert human evaluation.

\begin{table}[h!]
\centering
\caption{Automated evaluation of keyphrases on \texttt{FAIIR V1} classification task using the various matching methods. Sample-averaged scores are reported. The random baseline is from randomly predicting each label with 0.5 probability. \texttt{FAIIR V1} results are copied from \cite{faiirv1-2024}.}
\vspace{7mm}
\label{tab:kp_auto_results}
\begin{adjustbox}{width=0.75\textwidth}
\begin{tabular}{l|ccc}
Matching Method & Precision & Recall & F1-Score \\
\hline 
Random Baseline & 0.16 & 0.55 & 0.23 \\
\hline
Exact Matching & 0.45 & 0.24 & 0.30 \\
Exact Matching with Sublabels & 0.64 & 0.42 & 0.47 \\ 
Similarity with Labels (Threshold = 0.7) & 0.51 & 0.39 & 0.40 \\
Similarity with Labels and Unique Thresholds & 0.51 & 0.55 & 0.49 \\
Similarity with Sublabels (Threshold = 0.7) & 0.35 & 0.16 & 0.20 \\
Similarity with Sublabels and Unique Thresholds & 0.57 & 0.51 & 0.49 \\
\hline
\texttt{FAIIR V1} & 0.67 & 0.82 & 0.71 \\
\end{tabular}
\end{adjustbox}
\end{table}

\begin{figure*}[!tbp]
\centering

% ---------------- (a) ----------------
\begin{minipage}{\textwidth}
\centering
\subcaption*{\textbf{(a)}}
%\captionof{table}{Expert evaluation of the expanded issue taxonomy across four assessment dimensions.}
\vspace{5mm}
\label{tab:human_results_it}
\begin{adjustbox}{width=\textwidth}
\begin{tabular}{l|ccc}
Statement & Agree & Neutral & Disagree \\
\hline 
Overall, the new issue tags reflect the lived experiences of young people.
& \cellcolor{green!70}0.81 & \cellcolor{orange!30}0.17 & \cellcolor{red!25}0.03 \\
Overall, the new issue tags effectively identify emerging topics.
& \cellcolor{green!60}0.75 & \cellcolor{orange!35}0.19 & \cellcolor{red!35}0.06 \\
Overall, the new issue tags capture the subtle details of the conversations.
& \cellcolor{green!35}0.53 & \cellcolor{orange!55}0.33 & \cellcolor{red!65}0.14 \\
Overall, the new issue tags are useful for understanding the conversations.
& \cellcolor{green!45}0.61 & \cellcolor{orange!60}0.36 & \cellcolor{red!25}0.03 \\
\hline
Average & \cellcolor{green!50}\textbf{0.68} & \cellcolor{orange!45}\textbf{0.26}  &  \cellcolor{red!35}\textbf{0.065}\\
\end{tabular}
\end{adjustbox}
\end{minipage}

\vspace{6mm}

% ---------------- (b) ----------------
\begin{minipage}{\textwidth}
\centering
\subcaption*{\textbf{(b)}}
%\captionof{table}{Expert evaluation of keyphrase generative representation (KGR) across four assessment dimensions.}
\vspace{7mm}
\label{tab:kp_human_overall_results}
\begin{adjustbox}{width=\textwidth}
\begin{tabular}{l|ccc}
Statement & Agree & Neutral & Disagree \\
\hline 
Overall, the new key phrases reflect the lived experiences of young people.
& \cellcolor{green!70}0.74 & \cellcolor{orange!50}0.26 & \cellcolor{red!15}0.00 \\
Overall, the new key phrases effectively identify emerging topics in the conversations.
& \cellcolor{green!80}0.79 & \cellcolor{orange!35}0.18 & \cellcolor{red!25}0.03 \\
Overall, the new key phrases capture the subtle details of the conversations.
& \cellcolor{green!50}0.56 & \cellcolor{orange!60}0.36 & \cellcolor{red!50}0.08 \\
Overall, the new key phrases are useful for understanding the conversations.
& \cellcolor{green!75}0.75 & \cellcolor{orange!42}0.22 & \cellcolor{red!25}0.03 \\
\hline
Average & \cellcolor{green!68}\textbf{0.71} & \cellcolor{orange!47}\textbf{0.26} & \cellcolor{red!28}\textbf{0.04} \\
\end{tabular}
\end{adjustbox}
\end{minipage}

\vspace{6mm}

% ---------------- (c) ----------------
\begin{minipage}{\textwidth}
\centering
\subcaption*{\textbf{(c)}}
\includegraphics[width=16cm, height=7cm]{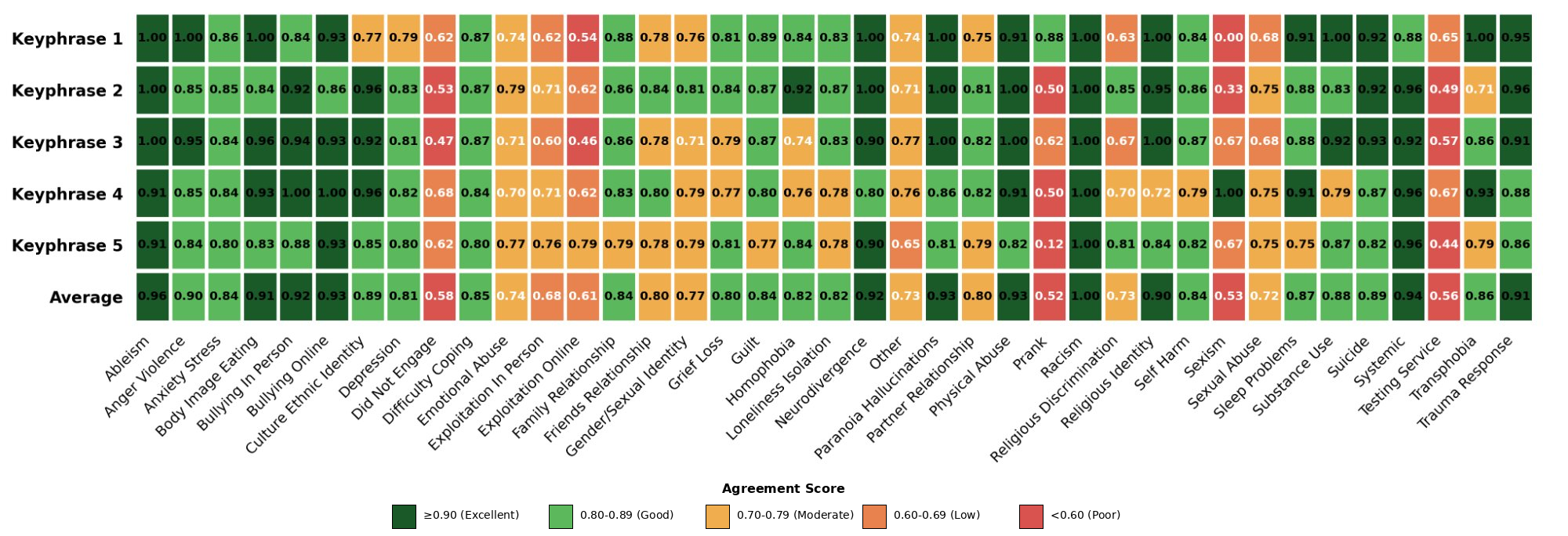}
%\captionof{figure}{Expert Agreement Ratios for KGR, divided vertically by the order, were generated (Keyphrase 1-5) and horizontally by the issue label assigned to the conversation. The Average column shows the mean agreement across all five keyphrase positions for each issue label.}
\label{fig:system2}
\end{minipage}

\caption{Expert evaluation of representation approaches for youth crisis conversations across three components. \textbf{(a)} Assessment of the expanded 39-issue taxonomy using expert judgments across four dimensions. \textbf{(b)} Evaluation of the KGR approach using the same expert rubric, allowing direct comparison with taxonomy-based issue tags. \textbf{(c)} Heatmap illustrating expert agreement ratios for generated keyphrases across issue labels and keyphrase positions (Keyphrase 1–5). The Average column reports the mean agreement across all keyphrase positions for each issue label.}
\label{fig:expert_evaluation_all}
\end{figure*}
%% Block 2 — Expert alignment (human evaluation)
We next assessed alignment between generated keyphrases and expert human judgment using the proposed KGR survey. Conversations were grouped according to expert-assigned issue tags using the Any aggregation scheme, and annotations were aggregated to compute the proportion of evaluators who judged a keyphrase as accurately reflecting the conversation topic, defined as the Expert Agreement Ratio shown in Figure~\ref{fig:expert_evaluation_all}(c). The expert review indicated that 81\% of generated keyphrases accurately reflected the conversation content. 

Agreement was highest for categories such as Racism (1.00), Ableism (0.96), and Suicide (0.94), and lowest for Sexism (0.53), Prank (0.53), Testing Service (0.56), Did Not Engage (0.58), and Exploitation, Online (0.61). Notably, categories within the same subdomain exhibited divergent performance; for example, Racism achieved perfect agreement, whereas Sexism showed substantially lower agreement. Lower-performing categories were concentrated within the "Other" group, suggesting greater difficulty distinguishing between conversation-specific content and general service-related exchanges.

Beyond agreement metrics, evaluators provided an assessment of KGR's utility shown in Figure~\ref {fig:expert_evaluation_all}(b). The majority of respondents (74\%) indicated that the generated keyphrases collectively covered the most relevant topics within conversations, with 71\% of responses positive, 26\% neutral, and 4\% negative across evaluation dimensions. Compared with the prior FAIIR V1 system, 24\% of respondents rated the keyphrases as much more effective, and 52\% as more effective.

\subsection{Emergent Themes Beyond the Fixed Taxonomy}

In addition, KGR surfaced several identity-related and situational themes that were not represented in the fixed 39-label taxonomy. These included identity and social-context expressions such as (1) residential school survivors, (2) Indigenous community contexts, (3) spiritual or religious trauma, (4) immigration-related challenges, (5) cultural conflict, (6) internalized racism, and (7) experiences of microaggressions. KGR also captured unique situational themes, including (1) car accident trauma, (2) runaway youth, (3) government funding stress, (4) forced marriage, (5) parental discovery of virginity loss, and (6) online addiction. These examples illustrate how generative representations can reveal culturally grounded and context-specific experiences that extend beyond predefined categorical labels, as shown in Figure \ref{tab:missed_examples}.

\begin{table}[h]
\centering
\caption{\textbf{Identity and situational themes recovered by keyphrase generation but absent from the original label taxonomy.}}
\label{tab:missed_examples}
\begin{tabular}{@{}p{0.28\linewidth} p{0.64\linewidth}@{}}
\toprule
\textbf{Category} & \textbf{Example keyphrases} \\
\midrule
Identity and social contexts &
(1) Residential school survivors; (2) Indigenous community; (3) spiritual crisis / religious trauma; (4) immigration-related challenges; (5) cultural differences / cultural disrespect; (6) internalised racism; (7) microaggressions. \\
\addlinespace
\midrule
Unique situational themes &
(1) Car-accident trauma; (2) runaway youth; (3) government funding issues; (4) forced marriage; (5) virginity loss / parental discovery; (6) online addiction; (7) caregiver burden. \\
\bottomrule
\end{tabular}
\end{table}

\subsection{Operational Application: Keyphrase-Based Topic Retrieval}
\label{sec:results_topic_retrieval}

Beyond surfacing emergent themes, we examined whether KGR can support a concrete operational workflow that fixed labels cannot. A common task at KHP is identifying conversations on report- or funding-application-specific topics that fall outside the predefined taxonomy. The existing workflow relies on a data analyst constructing keyword-based search queries~--~a process that is labour-intensive and sensitive to keyword choice. We compared keyphrase-based retrieval against this manual workflow on two representative topics: \textit{bullying} and \textit{body image}.

KGR retrieval substantially outperformed the manual workflow on both topics (Table~\ref{tab:insights}). For bullying, KGR achieved 70\% accuracy (F1~=~0.82) compared with 25\% (F1~=~0.40) for the analyst~--~a 45-percentage-point gap, with 14 of 20 conversations correctly identified versus 5 of 20. For body image, KGR achieved 73\% accuracy (F1~=~0.85) versus 53\% (F1~=~0.70), correctly identifying 11 of 15 conversations versus 8. Recall was 100\% for both methods on both topics, indicating that the performance difference was driven entirely by precision: KGR avoided the false positives that keyword search produced. These results show that keyphrase-based retrieval can reduce the manual effort required to assemble topic-specific reports while improving precision, supporting a concrete operational role for KGR that fixed-label classification cannot fill.

\begin{table}[!t]
\centering
\caption{Performance of KGR-based topic retrieval versus the current manual data-analyst workflow on two representative topics. $\Delta$ reports the absolute improvement of KGR over the analyst. KGR retrieval used cosine similarity between the target topic and generated keyphrases, with chain-of-thought prompting of Llama~3 for excerpt extraction. Expert review provided ground-truth relevance labels.}
\label{tab:insights}
\small
\renewcommand{\arraystretch}{1.2}
\begin{tabular}{l ccc c ccc}
\toprule
 & \multicolumn{3}{c}{\textbf{Bullying}} & & \multicolumn{3}{c}{\textbf{Body image}} \\
\cmidrule(lr){2-4} \cmidrule(lr){6-8}
\textbf{Metric} & KGR & Analyst & $\Delta$ & & KGR & Analyst & $\Delta$ \\
\midrule
Accuracy        & \cellcolor{green!12}\textbf{0.70} & 0.25 & \cellcolor{green!20}\textbf{+0.45} & & \cellcolor{green!12}\textbf{0.73} & 0.53 & \cellcolor{green!20}\textbf{+0.20} \\
Precision       & \cellcolor{green!12}\textbf{0.70} & 0.25 & \cellcolor{green!20}\textbf{+0.45} & & \cellcolor{green!12}\textbf{0.73} & 0.53 & \cellcolor{green!20}\textbf{+0.20} \\
Recall          & 1.00 & 1.00 & 0.00 & & 1.00 & 1.00 & 0.00 \\
F1-score        & \cellcolor{green!12}\textbf{0.82} & 0.40 & \cellcolor{green!20}\textbf{+0.42} & & \cellcolor{green!12}\textbf{0.85} & 0.70 & \cellcolor{green!20}\textbf{+0.15} \\
\midrule
Correct matches & \textbf{14\,/\,20} & 5\,/\,20 & \textbf{+9} & & \textbf{11\,/\,15} & 8\,/\,15 & \textbf{+3} \\
\bottomrule
\end{tabular}
\end{table}

\section{Discussion}

In this large-scale observational study of youth crisis conversations, taxonomy expansion improved expert agreement and topic coverage while maintaining high classification accuracy under consensus-based aggregation. These results confirm that structured multi-label classification remains an effective and scalable method for detecting high-priority clinical topics in crisis service data \cite{Graham2019AIMentalHealthOverview, Rose2007LabelsStigma,cmd1-npjdm-2023,broadbent-2023-frontiers-crisis-suicide}. At the same time, expert evaluation revealed difficulty capturing subtle contextual signals and evolving expressions of lived experience, even after substantial refinement of the taxonomy. Our findings suggest that improvements in classification performance do not necessarily resolve representational limitations in systems that rely exclusively on predefined labels.

The results point to a structural constraint inherent to taxonomy-based representations. Even when expanded, predefined label sets encode distress within fixed categorical boundaries. Increasing label granularity can reduce some blind spots, but cannot fully accommodate the diversity and evolution of language used in youth crisis conversations. Maintaining an exhaustively updated taxonomy becomes increasingly complex as sociocultural context and youth discourse change over time \cite{Moses2009SelfLabeling,gureje-2020-icd11-culture,lewisfernandez-2019-cultural-concepts,stupinski-2022-jmir-twitter-language}. Our findings therefore highlight an important distinction: high classification accuracy within a label space does not guarantee adequacy of the label space itself.

KGR addresses this limitation by reframing post-conversation issue identification as a representational problem rather than solely a classification task. KGR adds a complementary interpretive layer that preserves the auditability of taxonomy-based classification while surfacing contextual signals that fixed label systems cannot encode. Automated mapping demonstrated meaningful alignment between generated keyphrases and expert-derived labels, supporting construct validity. More importantly, expert review showed that KGR revealed identity-linked, culturally grounded, and situational themes absent from the expanded taxonomy, including experiences tied to social context and specific life circumstances \cite{faiirv1-2024}. The contribution is therefore not merely an incremental improvement in performance, but a broader shift toward keyphrase generation rather than the label method in crisis service analytics. These patterns suggest that generative representations may function most effectively as a mechanism for contextual enrichment and topic discovery and not a replacement for structured classification \cite{dinakar-2015-iui-fathom,althoff-2016-tacl-counseling,adhikary-2024-jmir-counseling-summarization,so-2024-jmir-formres-psychiatric-interview}.

Agreement also varied substantially across issue categories. KGR showed particularly strong alignment for high-salience categories such as racism, ableism, and suicide, while lower agreement was observed for categories such as sexism, prank, or testing service. These differences suggest that generative representations may perform best when distress is expressed through explicit semantic signals, while interactional or service-state categories may remain better captured by structured labels. Automated evaluation further demonstrated that the usefulness of generative representations depends strongly on how they are operationalized. Alignment performance varied substantially across mapping strategies, with poorly calibrated similarity-based methods performing worse than the random baseline. This finding suggests that generative representations require careful calibration to translate open-ended outputs into reliable analytic signals.

From an operational perspective, this layered representation may be particularly valuable for crisis service analytics. Post-conversation representations are used for supervision, triage auditing, quality assurance, and organizational reporting. Structured labels provide auditability and longitudinal comparability, both of which are essential for institutional oversight \cite{who-icd,snomed-intended-use,samhsa-2025-crisis-guidelines}. Generative descriptors can complement this infrastructure by surfacing contextual signals that would otherwise remain latent in narrative text. Anchoring generative outputs within a taxonomy-based framework therefore preserves operational continuity while extending the interpretive capacity of crisis service data systems \cite{meystre-haug-2005-problem-list,zhan-2021-patterns-structuring-clinical-text}.

The connection between structured classification and contextual richness is not unique to crisis services in mental health. Similar challenges arise in electronic patient messaging systems, clinical documentation, and other narrative health data systems where institutions must balance standardized reporting with interpretive nuance \cite{nguyen-2026-jamanetworkopen-smt, sulieman-2017-jbi-portal-cnn, cronin-2017-ijmedinfo-rule-ml}. The hybrid framework evaluated here, static labels as anchors, generative keyphrases as augmentation, may therefore provide a generalizable template for domains requiring both standardized reporting and interpretive depth.

This study has some limitations to discuss. Expert evaluation was conducted on a stratified subset of conversations and may not reflect performance across the full distribution of interactions. Automated evaluation of KGR relied on proxy comparisons to legacy labels rather than ground-truth keyphrases. Performance varied across issue categories, particularly for conversations with limited content or service-related exchanges. Finally, the findings are based on a retrospective analysis; a prospective evaluation is needed to assess real-world utility and safety. Future work should evaluate KGR in both prospective and operational settings, including responder-facing interfaces and decision-support workflows. Additional directions include improving robustness for ambiguous or low-information conversations, assessing temporal stability as youth language evolves, extending the approach to multilingual contexts, and using KGR outputs to inform principled taxonomy refinement and auditing for underrepresented themes.

\section{Ethics statement}

The study involving human participants was reviewed and approved by the Research Ethics Board at York University (certificate no. 2025-357). KHP is committed to the ethical and responsible use of data to enhance our services for youth, recognizing the importance of ethical principles in maintaining the trust of those we serve, especially the most vulnerable. This paper is aimed exclusively at applied research to improve service delivery and accessibility, with a special focus on the ethical application of AI to benefit our service network and frontline staff. Through this collaboration, we are committed to developing technological tools that offer a personalized and user-friendly experience for individuals seeking assistance. Upholding the privacy and confidentiality of our service users is paramount; we adhere to an ethical statement aligned with KHP’s privacy policy (https://kidshelpphone.ca/privacy-policy/), which includes a consent notice for research and rigorous data minimization. Our processes are transparent and accountable, and we comply with Canadian privacy regulations. We meticulously remove all direct identifiers from research data, adhering to industry standards for data anonymization, and securely store all research data within KHP’s infrastructure. This reflects our commitment to the highest standards of data security and ethical practice. By prioritizing ethical data use, KHP can leverage research to improve our services and deliver the best possible support for youth across Canada, embodying our commitment to integrity, respect, and responsibility in every action we take.

% =====================================================================
\section*{Data availability}
\addcontentsline{toc}{section}{Data availability}

The data supporting this study are not publicly available due to sensitivity concerns and data privacy following Kids Help Phone Privacy protocols. These data are stored in controlled-access storage at Kids Help Phone. Please contact \texttt{appliedresearch@kidshelpphone.ca} for more information.

% =====================================================================
\section*{Code availability}
\addcontentsline{toc}{section}{Code availability}

The code supporting this study is not publicly available due to sensitivity and data privacy concerns. Full code is available in a \href{https://github.com/KidsHelpPhone/AI-ML/tree/main/FAIIR_V2}{private GitHub} repository upon request. The repository also contains the code used to generate and process our data, including all parameters and settings. Please contact \texttt{appliedresearch@kidshelpphone.ca} for more information.

\section{Acknowledgment}

Kids Help Phone gratefully acknowledges the support from its community of accelerator Catalyst Partners, including The Lang Family Foundation and visionary investors who made this work possible (\url{https://kidshelpphone.ca/Catalyst_Partners}). The research was undertaken, in part, thanks to funding from the Natural Sciences and Engineering Research Council of Canada (NSERC) Discovery Grant and the Connected Minds Program, supported by the Canada First Research Excellence Fund (Grant CFREF2022-00010). The resources used in preparing this research were provided, in part, by the Province of Ontario, the Government of Canada through CIFAR, and companies sponsoring the Vector Institute (\url{https://www.vectorinstitute.ai/partnerships/}).

%Conclusions: In summary, our study demonstrates that a large language model–based assistant can effectively analyze and distill text-based counseling conversations, providing both structured classifications of issues and human-readable summaries or insights. By validating FAIIR V2’s performance on real-world youth crisis data and incorporating feedback from frontline clinicians, we have taken an early but significant step toward integrating AI safely into mental health practice. The system’s ability to highlight critical information and trends has the potential to enhance clinical decision-making, support quality improvement, and unlock new research insights from the vast troves of digital mental health communications that currently go under-utilized. At the same time, our work reinforces that human-centered design and rigorous evaluation are indispensable — the true value of AI in this domain will be realized only if it is developed and deployed in partnership with the people it aims to help (both providers and clients). Ongoing interdisciplinary efforts will focus on refining the technology, addressing its limitations, and ensuring ethical guardrails, so that tools like FAIIR V2 can ultimately be trusted in practice. With careful implementation, we believe that LLM-driven assistants can become an integral support for crisis responders, scaling up the reach of services and freeing human experts to do what they do best — providing empathy, understanding, and healing to those in need.

% =====================================================================
\section*{Author contributions}
\addcontentsline{toc}{section}{Author contributions}

A.B. and W.A. led the study conception, methodology, experimentation, analysis, and manuscript preparation. W.A. contributed to model development, automated evaluation, and code implementation. L.S. and J.R. led the expert evaluation framework, recruited Crisis Responders for annotation, and contributed clinical expertise on crisis service operations at Kids Help Phone. M.N. contributed to data infrastructure, annotation coordination, and operational integration at Vector Institute. E.D. supervised the project, provided methodological guidance, and contributed to study design and manuscript preparation. All authors reviewed and approved the final manuscript.

% =====================================================================
\section*{Competing interests}
\addcontentsline{toc}{section}{Competing interests}

The authors declare no competing interests.

\bibliographystyle{naturemag}
\bibliography{custom}
    
\appendix

\end{document}

\onehalfspacing

% ── Title page ──────────────────────────────────────────────
\begin{center}
  {\LARGE\bfseries SUPPLEMENTARY MATERIALS}\\[1em]
\end{center}

% ── Table of Contents ────────────────────────────────────────
\tableofcontents
\clearpage

% ============================================================
\appendix
% ============================================================

\section{Detailed Evaluation Framework and Mapping Strategies}
\label{sec:app_eval}

This appendix provides additional details on the evaluation framework used in the study and the automated mapping strategies applied to the KGR outputs.

\subsection{Alternate Label Distributions}
\label{sec:app_label_dists}

\begin{figure*}[!tbp]
 \centering
 \begin{subfigure}{1\linewidth}
   \centering
   \includegraphics[width=18 cm, height = 10 cm]{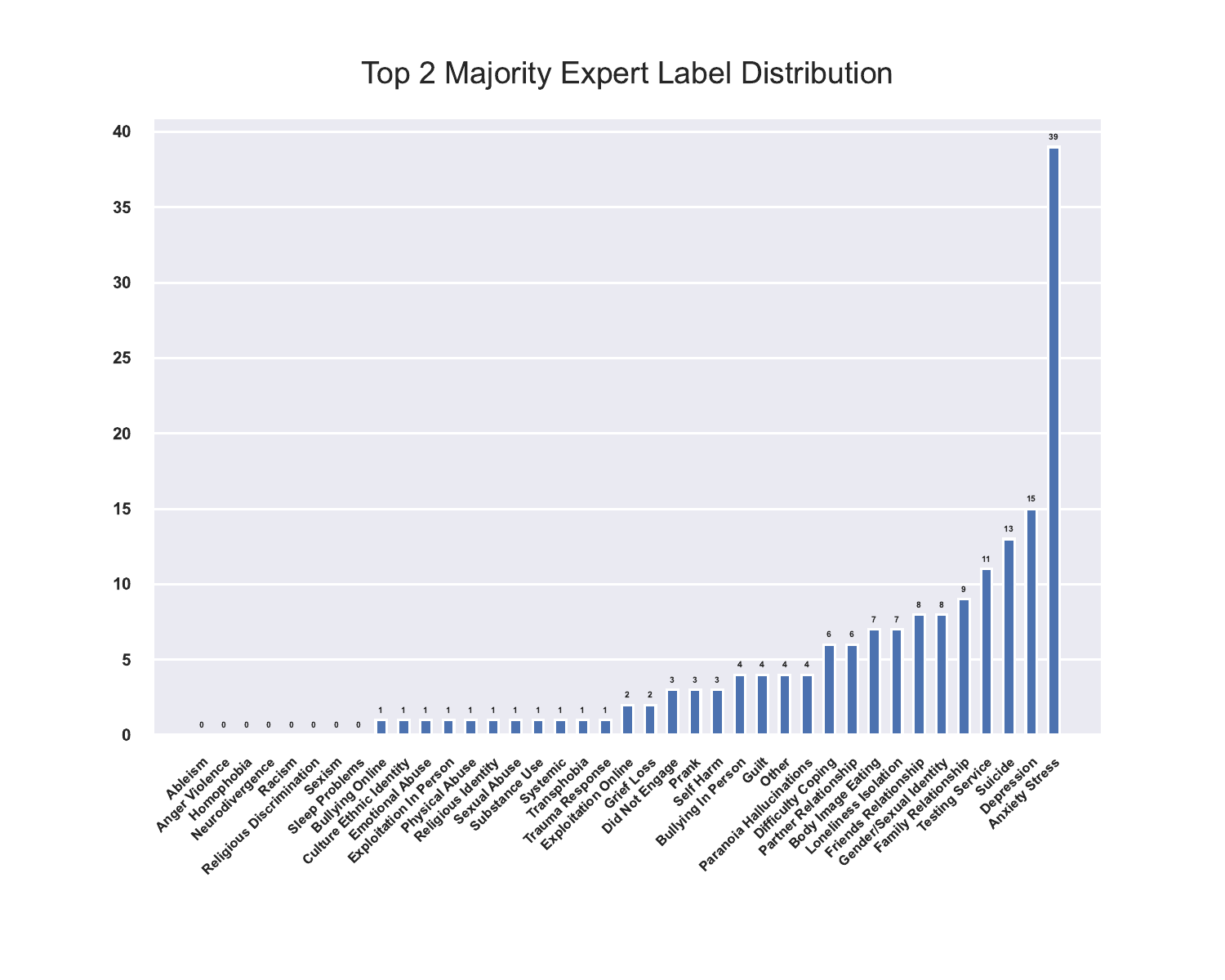}
   \caption{Top 2 Majority Expert Label Distribution}
   \label{fig:label_dist_top2_maj}
 \end{subfigure}

 \begin{subfigure}{1\linewidth}
   \centering
   \includegraphics[width=18 cm, height = 10 cm]{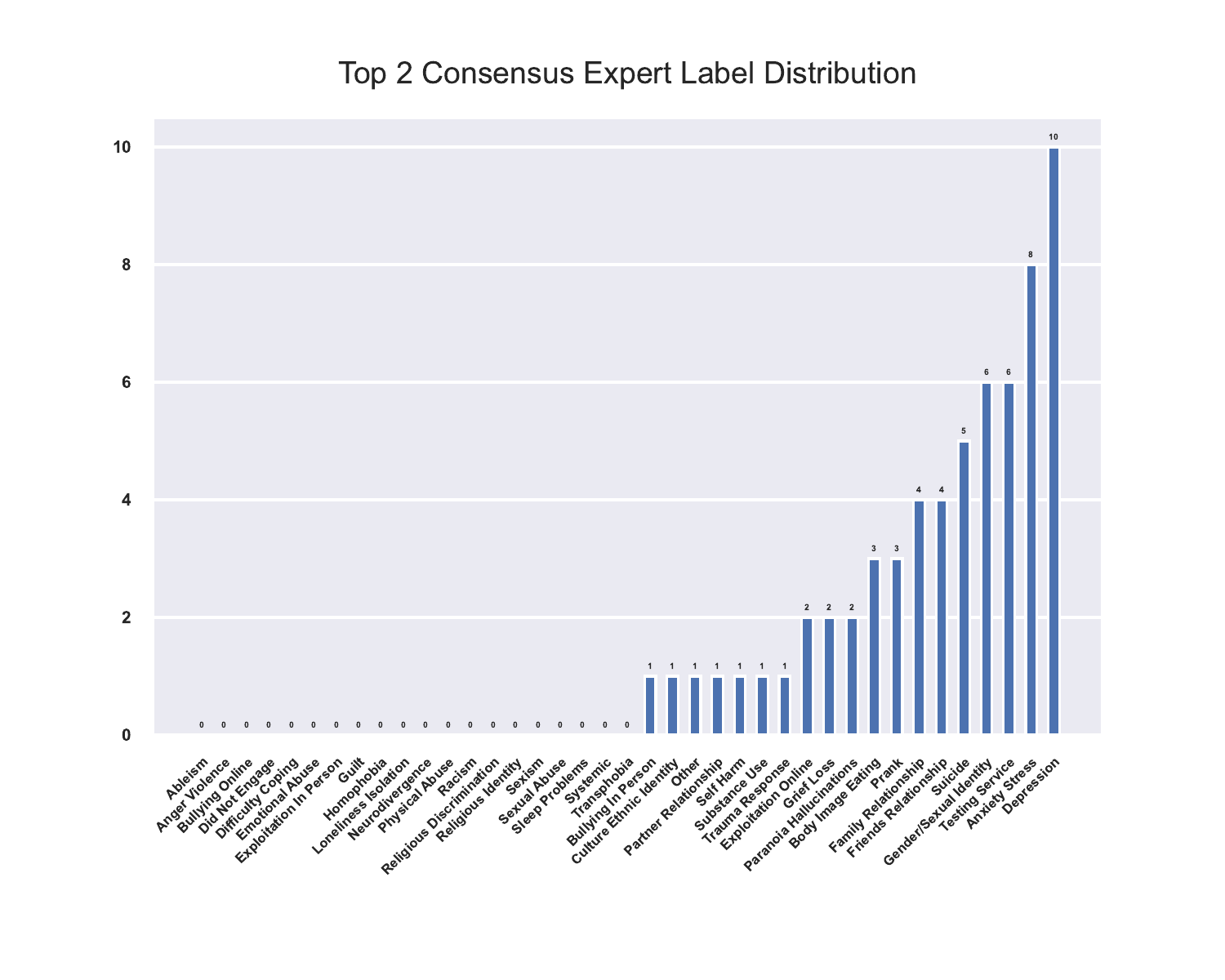}
   \caption{Top 2 Consensus Expert Label Distribution}
   \label{fig:label_dist_top2_consensus}
 \end{subfigure}
 
 \caption{Label distribution of 129 conversations annotated for the FAIIR V2 survey using (a) the Top 2 Majority assignment scheme and (b) the Top 2 Consensus assignment scheme.}
 \label{fig:label_dist_top2}
\end{figure*}

The label distributions presented in Figure \ref{fig:label_dist} are for the Any assignment scheme. The corresponding figures for the Top 2 Majority and Top 2 Consensus variants in our results are in Figure \ref{fig:label_dist_top2_maj} and \ref{fig:label_dist_top2_consensus}, respectively.

\subsection{Expert Evaluation Framework}

Both the taxonomy-based multi-label classification system and the proposed KGR approach were evaluated using a structured expert assessment framework that reflects real-world operational needs in crisis services. The framework was developed in collaboration with trained CRs at Kids Help Phone and focuses on how well each representation supports the interpretation of youth crisis conversations.

As summarized in Table~\ref{tab:issue_tag_evaluation}, the evaluation consisted of several components. First, experts assessed coverage, indicating whether the representation captured all relevant issues discussed in the conversation using a binary Yes/No response. Second, evaluators rated overall effectiveness across four dimensions using a three-point scale (Agree, Neutral, Disagree): reflection of lived experiences, identification of emerging topics, capture of subtle conversational details, and usefulness for understanding the interaction. These dimensions were selected to assess whether the representation aligns with the interpretive and supervisory needs of frontline responders.

Experts also performed a comparative assessment in which each representation was evaluated relative to the legacy 19-label \texttt{FAIIR V1} taxonomy \cite{faiirv1-2024,khp}. Responses were recorded on a five-point scale ranging from ``much less effective'' to ``much more effective.'' Finally, respondents were given the option to provide qualitative commentary to highlight strengths, limitations, or notable observations regarding the representation.

\subsection{Keyphrase-to-Label Mapping Strategies}

To enable automated evaluation of KGR outputs, generated keyphrases were mapped to the existing issue taxonomy using a set of progressively more flexible alignment strategies. These strategies are summarized in Table~\ref{tab:mapping_strategies}. The simplest method, Exact Matching, checks whether a generated keyphrase contains the exact name of a taxonomy label. To improve robustness to lexical variation, the Exact + Sublabels strategy expands each label with semantically related terms (e.g., \textit{Addiction} as a sublabel for \textit{Substance Abuse}). 

More flexible methods rely on embedding-based semantic similarity. In these approaches, both keyphrases and labels are embedded using the \texttt{all-MiniLM-L6-v2} sentence transformer model, and cosine similarity is computed to determine alignment. A match is recorded when similarity exceeds a predefined threshold. Because different labels exhibit different semantic variability, the Similarity + Unique Thresholds strategy introduces per-class threshold tuning to balance precision and recall.

Finally, the Similarity + Sublabels and Similarity + Sublabels + Unique Thresholds strategies combine semantic similarity with expanded label vocabularies, averaging similarity scores across multiple sublabels and optionally applying label-specific thresholds. These progressively more flexible strategies allow automated evaluation to capture semantic equivalence between generated keyphrases and taxonomy labels even when wording differs.

\begin{table}[h]
\centering
\caption{Evaluation Criteria Applied to both Multi-label Classification and Keyphrases Generative Representation (KGR) Approaches.}
\label{tab:issue_tag_evaluation}
\begin{tabular}{lp{0.65\textwidth}}
\hline
\textbf{Evaluation Component} & \textbf{Details} \\
\hline
Coverage & Yes/No response to whether the approach covered all relevant topics in conversations \\
\hline
\multicolumn{2}{l}{\textit{Overall Effectiveness (Agree/Neutral/Disagree):}} \\
\quad Lived Experiences & The approach reflects the lived experiences of young people \\
\quad Emerging Topics & The approach effectively identifies emerging conversation topics  \\
\quad Subtle Details & The approach captures the subtle details of the conversations \\
\quad Usefulness & The approach is useful for understanding the conversations \\
\hline
Comparative Assessment & Five-point scale comparing to 19-tag \texttt{FAIIR V1} system \\
\hline
Qualitative Input & Optional commentary on the approach \\
\hline
\end{tabular}
\end{table}

\begin{table}[h]
\centering
\caption{Keyphrase-to-label mapping strategies for issue tag classification.}
\label{tab:mapping_strategies}
\small
\begin{tabular}{p{0.28\textwidth}p{0.68\textwidth}}
\hline
\textbf{Strategy} & \textbf{Description} \\
\hline
Exact Matching & Baseline in which the keyphrases contain exact label names \\
\hline
Exact + Sublabels & Match keyphrases against labels and related terms (e.g., \textit{Addiction} for \textit{Substance Abuse}) \\
\hline
Similarity & Embed keyphrases and labels with \texttt{all-MiniLM-L6-v2}; match if cosine similarity exceeds threshold \\
\hline
Similarity + Unique Thresholds & Per-class threshold tuning to balance precision and recall \\
\hline
Similarity + Sublabels & Average cosine scores between keyphrases and all sublabels per class \\
\hline
Similarity + Sublabels + Unique Thresholds & Per-class thresholds applied to sublabel-averaged similarity scores \\
\hline
\end{tabular}
\end{table}
\newpage

\section{Experiment Set-Up}
\label{sec:set-up}
All experiments were performed on an AWS EC2 g5.12xlarge instance equipped with 4 Nvidia
A40 GPUs with the 8B Instruct version of LLaMA 3.0. Due to the faster nature of zero-shot
Testing, we didn't set up batching to fully utilize GPU memory. However, limiting the hardware
to just one GPU had insufficient memory for the length of conversations 6 plus generation output. Therefore, we recommend at least 40GB of RAM if the systems are moved to other locations. Label classification took 3.583 seconds per FAIIR V2 survey conversation and 4.850 seconds per conversation for keyphrase generation. Generation hyperparameters were kept to Hugging Face defaults, except we turned off sampling in favour of greedy decoding. The maximum number of tokens generated can also be limited to reduce memory requirements.

\section{Automated Evaluation of Keyphrase--Label Alignment}
\label{sec:app_auto_eval}

Table~\ref{tab:kp_auto_results} reports the automated evaluation results used to assess how well generated keyphrases align with the issue labels defined in the \texttt{FAIIR V1} taxonomy \cite{faiirv1-2024}. Because ground-truth keyphrases are unavailable, we treat the existing 19 survey labels as proxy references and evaluate the ability of generated keyphrases to recover these categories using different matching strategies. Several alignment methods were examined. The simplest method, \textit{Exact Matching}, assigns a label only when the generated keyphrase explicitly contains the label name. The \textit{Exact Matching with Sublabels} strategy extends this approach by including semantically related terms associated with each label. More flexible strategies rely on embedding-based semantic similarity using the \texttt{all-MiniLM-L6-v2} model. In these approaches, keyphrases and labels are embedded in a shared vector space, and cosine similarity is used to determine alignment when the cosine score exceeds a predefined threshold.

To improve robustness across heterogeneous issue categories, two additional variants introduce label-specific thresholds. These \textit{Unique Threshold} strategies allow different similarity cutoffs for different issue classes, helping balance precision and recall. Additional variants also incorporate sublabels when computing similarity scores, averaging similarity across multiple label-related terms. Performance is reported using sample-averaged precision, recall, and F1-score. The random baseline assigns each label a probability of 0.5 and provides a lower bound for comparison. Results demonstrate that incorporating sublabels and class-specific thresholds substantially improves alignment between generated keyphrases and existing taxonomy labels, indicating that generative outputs capture semantically meaningful issue representations.

\begin{table}[h!]
\centering
\caption{Automated evaluation of keyphrases on \texttt{FAIIR V1} classification task using the various matching methods proposed. Sample-averaged scores are reported. The random baseline is from randomly predicting each label with 0.5 probability. \texttt{FAIIR V1} results are copied from \citet{faiirv1-2024}.}
\vspace{7mm}
\label{tab:kp_auto_results}
\begin{adjustbox}{width=0.75\textwidth}
\begin{tabular}{l|ccc}
Matching Method & Precision & Recall & F1-Score \\
\hline 
Random Baseline & 0.16 & 0.55 & 0.23 \\
\hline
Exact Matching & 0.45 & 0.24 & 0.30 \\
Exact Matching with Sublabels & 0.64 & 0.42 & 0.47 \\ 
Similarity with Labels (Threshold = 0.7) & 0.51 & 0.39 & 0.40 \\
Similarity with Labels and Unique Thresholds & 0.51 & 0.55 & 0.49 \\
Similarity with Sublabels (Threshold = 0.7) & 0.35 & 0.16 & 0.20 \\
Similarity with Sublabels and Unique Thresholds & 0.57 & 0.51 & 0.49 \\
\hline
\texttt{FAIIR V1} & 0.67 & 0.82 & 0.71 \\
\end{tabular}
\end{adjustbox}
\end{table}

\section{List of Sublabels and Thresholds for Automatic \\ Keyphrase Evaluation}

The sublabels chosen for the original 19 issue tags are given in Table \ref{tab:sublabels}. They were used for automatic keyphrase evaluation. The single thresholds were used for the ``Similarity with Labels and Unique Thresholds'' row and the multi-thresholds were used for ``Similarity with Sublabels and Unique Thresholds''.

\begin{table}[ht!]
\centering
\caption{List of sublabels for keyphrase automatic evaluation on \texttt{FAIIR V1} Survey. The single threshold is used for similarity scores comparing only with the main label name while the multi-threshold is used for the average of sublabels.}
\vspace{7mm}
\label{tab:sublabels}
\begin{adjustbox}{width=\textwidth}
\begin{tabular}{r|l|c|c}
Issue Label & Issue Sublabel(s) & Single-Threshold & Multi-Threshold \\ 
\hline
3rd Party & Third Party, On behalf of & 0.5 &0.5 \\ 
Abuse, emotional & Emotional Abuse, Verbal Abuse & 0.75& 0.65 \\ 
Abuse, physical & Physical Abuse, Beat, Hit & 0.75 &0.55 \\ 
Abuse, sexual & Sexual Abuse, Rape, Harass, Consent &0.7 &0.7 \\ 
Anxiety/Stress & Anxiety, Stress, Distress, Fear, Panic, Scared, Uncertain, Overwhelm, Pressure & 0.55 &0.43 \\ 
Bully & Cyberbully, Judged & 0.6 &0.5 \\ 
DNE & Did not engage, Unresponsive &0.5 &0.5 \\ 
Depressed & Sad, Despair, Hopeless, Feeling Down, Feeling Low, Lack of Motivation, Negative & 0.6 &0.43 \\ 
Eating Body Image & Eating Disorder, Disordered Eating, Body Dysmorphia, Body Image, Weight, Fat, Anorexia, Bulimia & 0.3 &0.4 \\ 
Gender/Sexual Identity & Sexual Identity, Gender, Gay, Lesbian, Queer, Bi, Trans, Transgender, Non-Binary, Two-Spirit, Dysphoria & 0.4 &0.4 \\ 
Grief & Loss, Lost, Passed & 0.6 &0.5 \\ 
Isolated & Alone, Helpless & 0.4 &0.45 \\ 
Other & Unsure, Not Applicable, NA & 0.5 &0.5 \\ 
Prank & Vulgar, Joke &0.5 &0.5 \\ 
\multirow{2}*{Relationship} & Mom, Dad, Mother, Father, Parental, Care-Giver, Sister, Brother, Sibling, Aunt, Uncle, & \multirow{2}*{0.4} & \multirow{2}*{0.33} \\
& Cousin, Grandparent, Grandma, Grandpa, Partner, Boyfriend, Girlfriend, Friend, Family & & \\ 
Self Harm & & 0.75 &0.75 \\ 
Substance Abuse & Addiction, Dependent, Relapse, Alcohol, Drugs, Rehab & 0.65 &0.45 \\ 
Suicide & Kill Self, Suicidal Ideation, End Life, Suicidal Thoughts & 0.65 &0.65 \\ 
Testing & Practice, Information Seeking, Checking & 0.5&0.5\\ 
\end{tabular}
\end{adjustbox}
\end{table}

\section{Human Evaluation Survey Odd Cases}

\label{sec:oddcases}

During the human evaluation survey, a small number of conversations showed unusual or inconsistent patterns between the model's predicted labels and expert annotations. Below, we present three illustrative cases where the annotations highlight ambiguity, disagreement, or edge cases in the taxonomy.

\subsection{Case 1: Conv 103}

\textit{Conversation:}

Service User: TEST. \\
Service User: I work with [scrubbed] residential school survivors is this something they could use?. \\
Crisis Responder: Thanks so much for your interest in the program. This is definitely a program that residential school survivors could use. \\
Crisis Responder: There is a great team of volunteers here ready to listen and support them. \\
Service User: This is fabulous news! Thank you. \\
Crisis Responder: You're very welcome!. \\
Service User: I will share the text in the event someone is in crisis good night. \\

\noindent\textit{Model-predicted labels:} 
\verb|['Did Not Engage', 'Systemic', 'Testing Service']|

\noindent The table below shows how experts distributed their ratings across related labels, revealing a mix of ``Systemic'', ``Testing Service'', and culture-related themes for a brief, largely informational exchange.

\begin{table}[h!]
\begin{adjustbox}{width=\textwidth}
\begin{tabular}{|l|l|l|l|l|r|r|}
\hline
Trauma Response                  & Emotional Abuse        & Sexual Abuse           & Physical Abuse         & Culture Ethnic Identity & \multicolumn{1}{l|}{Systemic} & \multicolumn{1}{l|}{Testing Service} \\ \hline
\textbf{}                        &                        &                        &                        & \multicolumn{1}{r|}{1}  & 2                             & 3                                    \\ \hline
\multicolumn{1}{|r|}{\textbf{}}  &                        &                        &                        &                         & 2                             & 1                                    \\ \hline
\multicolumn{1}{|r|}{\textbf{2}} & \multicolumn{1}{r|}{3} & \multicolumn{1}{r|}{5} & \multicolumn{1}{r|}{4} & \multicolumn{1}{r|}{8}  & 6                             & 1                                    \\ \hline
\end{tabular}
\end{adjustbox}
\caption{Expert annotation distribution for Conversation 103. Each row represents one annotator; values indicate ranking positions assigned to each label.}
\end{table}

\subsection{Case 2: Conv 124}

\noindent\textit{Original labels from the old survey:} \\
\verb|['Isolated', 'Relationship', 'Suicide']|

\noindent\textit{Current Model-predicted labels:} \\
\verb|['Depression', 'Relationship', 'Suicide']|

\noindent Here, the model largely aligns with the original labels (Relationship, Suicide) but replaces ``Isolated'' with ``Depression.'' The table shows how experts scored related emotional and risk categories, highlighting overlap between loneliness, guilt, and depressive themes.

\begin{table}[h!]
\begin{adjustbox}{width=\textwidth}
\begin{tabular}{|l|l|l|l|l|l|l|r|}
\hline
Suicide                 & Anger Violence        & Anxiety Stress         & Sleep Problems        & Depression             & Paranoia Hallucinations & Guilt                  & \multicolumn{1}{l|}{Loneliness Isolation} \\ \hline
                        &                       & \multicolumn{1}{r|}{2} & \multicolumn{1}{r|}{} &                        &                         & \multicolumn{1}{r|}{3} &                                           \\ \hline
\multicolumn{1}{|r|}{4} & \multicolumn{1}{r|}{} &                        &                       & \multicolumn{1}{r|}{2} & \multicolumn{1}{r|}{}   &                        & 3                                         \\ \hline
\multicolumn{1}{|r|}{1} & \multicolumn{1}{r|}{} & \multicolumn{1}{r|}{2} & \multicolumn{1}{r|}{} & \multicolumn{1}{r|}{3} & \multicolumn{1}{r|}{}   & \multicolumn{1}{r|}{4} &                                           \\ \hline
\end{tabular}
\end{adjustbox}
\caption{Expert annotation distribution for Conversation 124. Each row represents one annotator; values indicate ranking positions assigned to each label.}
\end{table}

\subsection{Case 3: Conv 105}

\noindent In this conversation, experts distribute their ratings across multiple overlapping emotional and relational categories, illustrating how complex presentations can lead to label dispersion and disagreements between annotators.

\begin{table}[h!]
\begin{adjustbox}{width=\textwidth}
\begin{tabular}{|r|r|r|r|r|l|l|r|}
\hline
\multicolumn{1}{|l|}{Anxiety Stress} & \multicolumn{1}{l|}{Sleep Problems} & \multicolumn{1}{l|}{Depression} & \multicolumn{1}{l|}{Loneliness Isolation} & \multicolumn{1}{l|}{Grief Loss} & Difficulty Coping      & Trauma Response        & \multicolumn{1}{l|}{Family Relationship} \\ \hline
1                                    &                                     & 2                               &                                           & 3                               & \multicolumn{1}{r|}{5} & \multicolumn{1}{r|}{}  & 4                                        \\ \hline
4                                    & 3                                   & 2                               & 1                                         &                                 &                        &                        & 5                                        \\ \hline
2                                    & 6                                   & 1                               &                                           & 4                               & \multicolumn{1}{r|}{3} & \multicolumn{1}{r|}{5} & 7                                        \\ \hline
\end{tabular}
\end{adjustbox}
\caption{Expert annotation distribution for Conversation 105. Each row represents one annotator; values indicate ranking positions assigned to each label.}
\end{table}

These examples illustrate ``odd'' or ambiguous cases where label assignment is challenging, and where both model predictions and expert annotations reveal the inherent complexity of youth crisis conversations.

\section{Case Study: Chain-of-Thought Prompting and Insights}
\label{appendixcot}
\subsection{Methodology}
\paragraph{Chain-of-Thought (CoT)} CoT zero-shot learning \citep{zero-shot-cot-2022} is a novel approach that enhances LLM performance and yields more insights. Zero-shot learning aims to overcome the limitation of having no training data available by enabling a model to perform tasks it has never encountered during its training, relying solely on its general understanding and the task description. CoT zero-shot learning further this concept by prompting the model to generate intermediate steps, or ``thoughts,'' as it works towards the solution. This process mimics human problem-solving by breaking a problem into smaller components, leading to more effective results.  This approach has shown promise in enhancing model performance across various domains, specifically in LLMs. In this work, we use CoT zero-shot learning to explain why the model predicts specific labels in the conversation. This helps the frontline workers understand the reasoning behind the generated labels and improve future model performance.

\paragraph{Insights Generation.} A primary motivator behind open-ended keyphrase generation was to enable the identification of new issues and topics not covered by a pre-identified classification set. As an example of how value can be derived from them, we propose \textit{Insights Generation}. For various reports and funding applications, KHP searches its database of conversations for instances of specific topics that often do not directly correspond to existing labels. The current workflow is rudimentary and involves substantial human engineering of SQL queries. As an alternative, Insights Generation measures similarity scores between a proposed topic (e.g. Climate Anxiety) and the keyphrases of each conversation in a selected time period. Optionally, for scores above a threshold, LLaMA 3 can be prompted with the conversation and asked to return verbatim quotes relevant to the topic at hand. Automating this procedure substantially reduces the burden on report authors, allowing them to focus on writing rather than data procurement.

For evaluation, we requested a data analyst at Kids Help Phone Company to identify conversations related to a specific topic from the survey list. The data analyst manually searches for the main keywords of the topic in these conversations and then matches the filtered conversation with the specific topic. On the other hand, we use the model insights generation to generate a list of conversations that match the topic. Regarding the truth labels, we use the expert evaluation to determine how many conversations align with a specific topic. We evaluate the two methods using metrics such as accuracy, precision, recall, and F1-score to compare their performance.

\subsection{Results}

Table \ref{tab:insights} compares the performance of the model and a data analyst in identifying conversations related to Bullying and Body Image. For the model, we provide a prompt with a question to identify a specific topic from a list of conversations, along with an explanation (CoT). We calculate accuracy, precision, recall, F1-score, and the number of matches and non-matches for evaluation. For the bullying topic, the model achieved 70\% accuracy, significantly higher than the data analyst's 25\%. This suggests that the model is more consistent in correctly identifying bullying-related conversations. 

The model and the data analyst have 70\% and 25\% precision, respectively. This indicates that when the model identifies a conversation as related to bullying, it is more likely to be correct than the data analyst. Both the model and the data analyst achieved 100\% recall, indicating they could identify all bullying-related conversations in the dataset. The model's F1-score is 82\%, and the data analyst's is 40\%, demonstrating that the model sustains a better balance between precision and recall. The model correctly identified 14 of 20 matches, compared to the data analyst, who identified only 5. 

For conversations about body image, the model's accuracy is 73\%, while the data analyst's is 53\%. This indicates that the model is more accurate at identifying conversations about body image. The model also outperformed the data analyst in precision, achieving a score of 73\% compared to 53\%. This suggests that the model is more effective at accurately identifying relevant conversations when making predictions. Both the model and the data analyst achieved a recall of 100\%, indicating that all conversations related to body image were correctly identified. The model achieved an F1-score of 85\%, which is higher than the data analyst's 70\%. This suggests that the model performs better in balancing precision and recall when identifying body image-related conversations. The model correctly identified 11 out of 15 matches, while the data analyst identified 8. 

\begin{table}[h!]
\centering
\caption{Model vs. Data Analyst Performance on Identifying Topics from the Conversations.}
\vspace{7mm}
\label{tab:insights}
\begin{adjustbox}{width=0.7\textwidth}
\begin{tabular}{|c|cc|cc|}
\hline
Topic                  & \multicolumn{2}{c|}{Bullying}                                  & \multicolumn{2}{c|}{Body Image}                                \\ \hline
\multicolumn{1}{|l|}{} & \multicolumn{1}{l|}{Llama 3 Model} & \multicolumn{1}{l|}{Data analyst} & \multicolumn{1}{l|}{Llama 3 Model} & \multicolumn{1}{l|}{Data analyst} \\ \hline
Accuracy               & \multicolumn{1}{c|}{70\%}  & 25\%                              & \multicolumn{1}{c|}{73\%}  & 53\%                              \\ \hline
Precision               & \multicolumn{1}{c|}{70\%}  & 25\%                              & \multicolumn{1}{c|}{73\%}  & 53\%                              \\ \hline
Recall                 & \multicolumn{1}{c|}{100\%} & 100\%                             & \multicolumn{1}{c|}{100\%} & 100\%                             \\ \hline
F1-score               & \multicolumn{1}{c|}{82\%}  & 40\%                              & \multicolumn{1}{c|}{85\%}  & 70\%                              \\ \hline
Match                  & \multicolumn{1}{c|}{14}    & 5                                 & \multicolumn{1}{c|}{11}    & 8                                 \\ \hline
No match               & \multicolumn{1}{c|}{6}     & 15                                & \multicolumn{1}{c|}{4}     & 7                                 \\ \hline
Total                  & \multicolumn{1}{c|}{20}    & 20                                & \multicolumn{1}{c|}{15}    & 15                                \\ \hline
\end{tabular}
\end{adjustbox}
\end{table}

The results indicate that the model consistently outperforms the data analyst across multiple evaluation metrics for both topics, Bullying and Body Image. The model demonstrates higher accuracy, precision, and F1-score in both cases, indicating greater reliability in identifying relevant conversadata analysts' reliance on manual effort between the model and the data analyst further emphasizes the model's advantage in maintaining a solid balance between identifying all relevant conversations (recall) and ensuring that those identified are indeed relevant (precision). In conclusion, using the model, particularly with the zero-shot learning approach, has proven to enhance performance in identifying relevant conversations within specific topics. This suggests that such models can be valuable tools for automating and improving the accuracy of topic identification in large datasets, thereby reducing the reliance of data analysts on manual efforts.

\bibliographystyle{unsrtnat}

\bibliography{custom}